\documentclass[english,twocolumn]{svjour3}
\usepackage[T1]{fontenc}
\usepackage[utf8]{inputenc}
\usepackage{babel}
\usepackage{array}
\usepackage{varioref}
\usepackage{prettyref}
\usepackage{textcomp}
\usepackage{url}
\usepackage{multirow}
\usepackage{amsmath}
\usepackage{graphicx}
\usepackage{nameref}

\makeatletter

\providecommand{\tabularnewline}{\\}

\RequirePackage{fix-cm}

\makeatletter
\g@addto@macro{\UrlBreaks}{\UrlOrds}
\makeatother

\smartqed  % flush right qed marks, e.g. at end of proof

\usepackage{hyperref}

\makeatother

\begin{document}
\title{Learning from similarity and information extraction from structured
documents\thanks{The work was supported by the grant SVV-2020-260583.}}
\author{Martin Hole\v{c}ek}
\institute{Martin Hole\v{c}ek \at Faculty of Mathematics and Physics, Charles
University, Department of Numerical Mathematics \\
Prague, Czech Republic\\
Tel.: +420-603775375\\
orcid.org/0000-0002-1008-1567\\
\email{martin.holecek.ai@gmail.com}}
\date{Received: date / Accepted: date}
\maketitle
\begin{abstract}
The automation of document processing is gaining recent attention
due to the great potential to reduce manual work through improved
methods and hardware. Any improvement of information extraction systems
or further reduction in their error rates has a significant impact
in the real world for any company working with business documents
as lowering the reliability on cost-heavy and error-prone human work
significantly improves the revenue. In this area, neural networks
have been applied before – even though they have been trained only
on relatively small datasets with hundreds of documents so far.

To successfully explore deep learning techniques and improve the information
extraction results, a dataset with more than twenty-five thousand
documents has been compiled, anonymized and is published as a part
of this work. We will expand our previous work where we proved that
convolutions, graph convolutions and self-attention can work together
and exploit all the information present in a structured document.
Taking the fully trainable method one step further, we will now design
and examine various approaches to using siamese networks, concepts
of similarity, one-shot learning and context/memory awareness. The
aim is to improve micro $F_{1}$ of per-word classification on the
huge real-world document dataset.

The results verify the hypothesis that trainable access to a similar
(yet still different) page together with its already known target
information improves the information extraction. Furthermore, the
experiments confirm that all proposed architecture parts (siamese
networks, employing class information, query-answer attention module
and skip connections to a similar page) are all required to beat the
previous results.

The best model improves the previous state-of-the-art results by an
$8.25\,\%$ gain in $F_{1}$ score. Qualitative analysis is provided
to verify that the new model performs better for all target classes.
Additionally, multiple structural observations about the causes of
the underperformance of some architectures are revealed.

All the source codes, parameters and implementation details are published
together with the dataset in the hope to push the research boundaries
since all the techniques used in this work are not problem-specific
and can be generalized for other tasks and contexts. 

\keywords{one-shot learning \and information extraction \and siamese networks
\and  similarity \and attention}

\end{abstract}

\ifdefined\dodeclarations

\section*{Declarations}

\subsection*{Funding}

The work was supported by the grant SVV-2020-260583. Partial financial
support was received from Rossum and Charles University.

\subsection*{Conflicts of interest/Competing interests}

The author (Martin Hole\v{c}ek) has received financial support from
Rossum and from Charles University – where he pursues a PhD. The author
is in an employment and/or contractual relationship with the following
subjects: Rossum, Medicalc and AMP Solar Group.

\subsection*{Availability of data and material (data transparency)}

The anonymized version of the dataset was made publicly available
at \cite{codeanddata}, together with all the codes. The improvement
of previous results can be reproduced even on the anonymized data
while not disclosing any sensitive information.

\subsection*{Code availability (software application or custom code)}

The source codes are made public in a Github repository \cite{codeanddata}.

\subsection*{Author's contributions}

The principal author is responsible for the study concept and design,
execution, coding and research. The rest of the Rossum team is responsible
for data acquisition, annotation and storage, and also for creating
a working product and environment, in which it was possible to do
a scientific study of this scope. 

\section*{Biographical summary}

\begin{minipage}[t][1.25in]{1in}%
\includegraphics[width=1in]{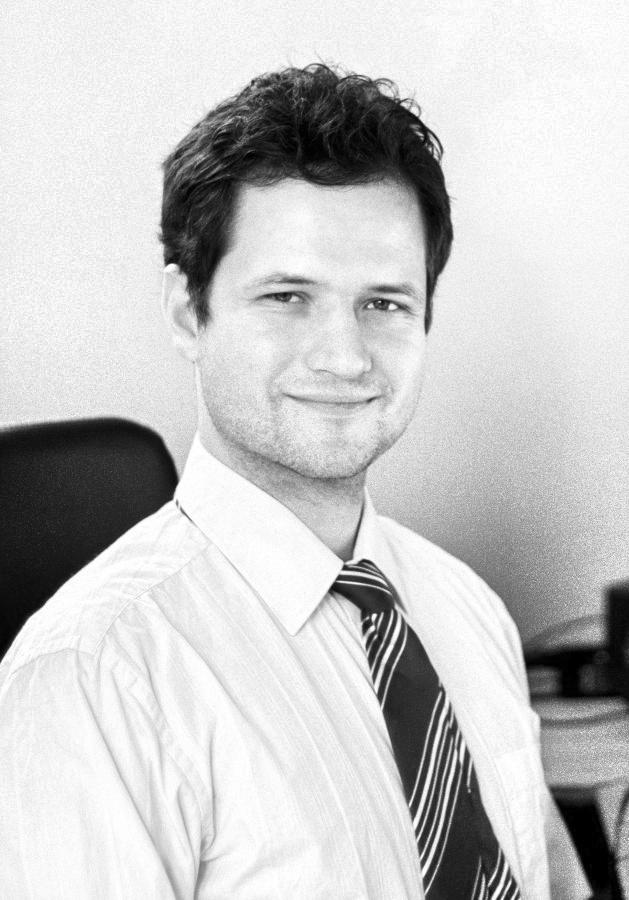}%
\end{minipage} 

Martin Hole\v{c}ek is a PhD student at the Department of Numerical
Mathematics, Faculty of Mathematics and Physics of Charles University
in Prague, with a masters degree in numerical mathematics. He has
been working in the field of document processing for two years in
Rossum.ai (see the previous work \cite{Ours}). His professional focus
is mostly deep neural networks.

\else \fi

\section{Introduction\label{sec:Introduction}}

Our goal is to improve information extraction from business documents
and contribute to the field of automated document processing. This
work leads to a higher success metric and enables less manual work
regarding data entry and/or annotation in the industry.

To put the work in context and define the terms closely let's briefly
recall the definition of the task, the motivation and add more details.

\paragraph{Information extraction task\label{par:Information-extraction-task}}

The general problem of information extraction is not a new problem
(see more works referenced in \prettyref{subsec:Related-works}).
A survey on information extraction methods \cite{Cowie1996InformationE}
defines the task as: ``Information Extraction starts with a collection
of texts, then transforms them into information that is more readily
digested and analyzed. It isolates relevant text fragments, extracts
relevant information from the fragments, and then pieces together
the targeted information in a coherent framework''.

The relevant collection of texts for this study are the texts in business
documents such as invoices, pro forma invoices and debit notes. The
targeted information is a classification of the texts that helps in
automating various business processes – such as automated payment
for invoices.

\paragraph{Motivation\label{par:Motivation}}

The typical user of our method would be any company medium-sized and
bigger because, at some point, companies start to spend significant
time on document processing. Details are harder to find in referenced
and peer-reviewed works since the companies keep their spending information
secret. Approximations from unofficial (non-scientific) sources as
\cite{costproc} and \cite{tenhunen2010assessing} lead to an estimate
of how a success metric translates to company savings. A typical medium-sized
company can have approximately $25\,000$ invoices per month and even
just $1\,\%$ improvement roughly translates to more than $500$ dollars
saving monthly and scales with the company size. Note that this is
just a heuristics and thus we do not define the metric exactly.

\paragraph{Details and overview\label{par:Details}}

As stated, we will focus on business documents. The explicit category
of the documents varies. Existing works on information extraction
\cite{RibaTableDetect,kosala10.1007/3-540-45681-3_25,Smith97informationextraction,liu2019graph}
define these as ``visually rich documents'', ``structured'', or
``semi-structured''. 

We will use the name ``structured documents'' throughout this work
since the structure of the documents is clear and understandable to
a human working in relevant fields, even though the specific structure
varies. Moreover, the documents are machine-readable up to the detail
of individual words and pictures (incl. their positions) on a page,
but for a machine, they are not ``understandable'' with respect
to the goal of important information extraction.

It is important to classify all of the information that is needed
in the financial/accounting industry, for the ``users'' of the documents.
For example, the payment details, amount to be paid, issuer information
etc. The input is a document's page and the goal is to identify and
output all of the words and entities in the document that are considered
important, along with their respective classifications.

\begin{figure}
\includegraphics[width=1\columnwidth]{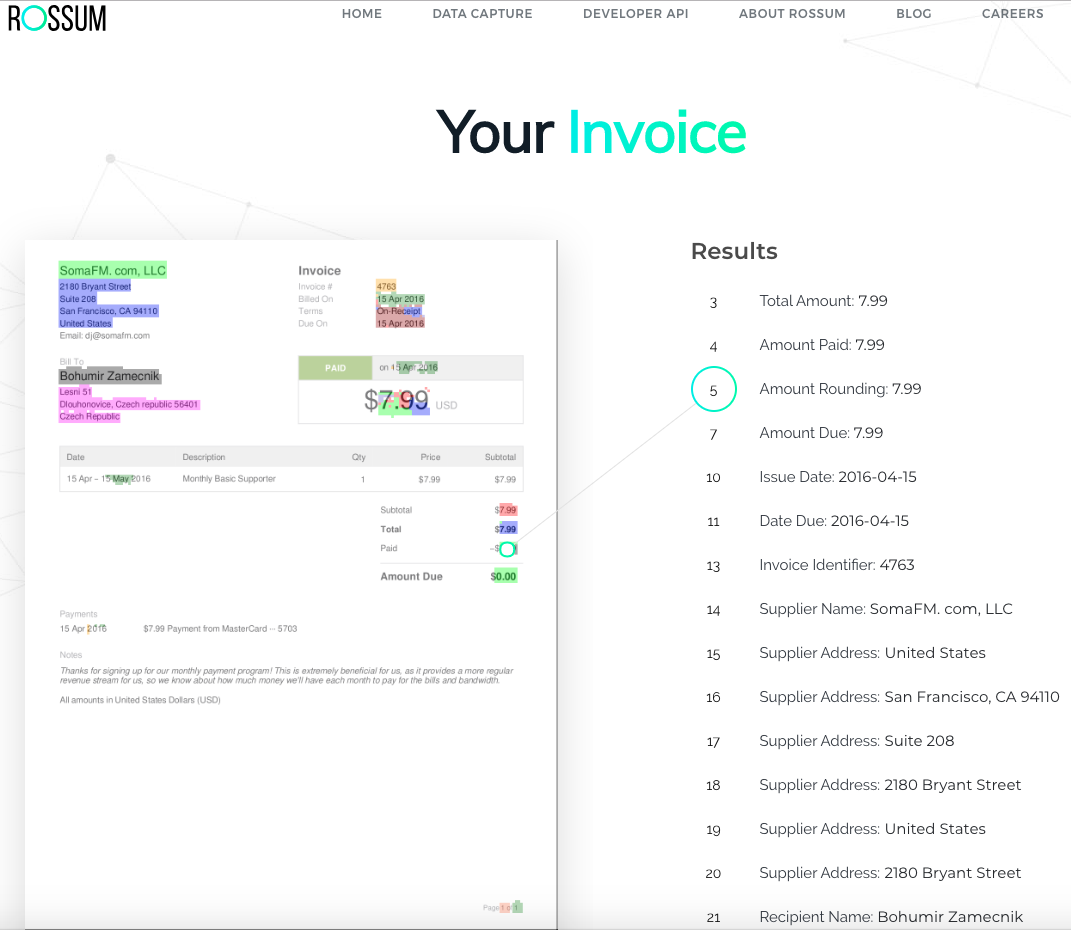}

\caption{\label{fig:Example}Example of an invoice and an extraction system
together with its output. This example should also illustrate why
invoices are called “structured documents”. We can see that when the
various information contained in the document is visually grouped
together, it usually belongs together. There is a heading ’Invoice’
under which segments of information about the invoice are written
next to their explanations. Some rectangular areas do not have these
explanations, and to find out what rectangular area speaks about the
sender and supplier, one has to look for a small ’Bill To:’ heading.
Please note, these specific rules apply only to this example and other
invoices are notably different. (Online image source \cite{invoiceexample}).}
\end{figure}

One example of an input invoice and output extraction can be seen
in \prettyref{fig:Example}. As you can see, the documents are not
easily understandable inputs. An example of trivial inputs would be
an XML document that has the desired target classes incorporated in
a machine-readable way.

With this study, we aim to expand previous work (\cite{Ours}, also
referenced as “previous”), in which we have already shown that neural
networks can succeed in the task of extracting important information
and even identifying whole, highly specific tables.

As argued before, every improvement matters and so in this work, the
focus is on improving the metrics by selecting relevant techniques
from the deep learning field. A classical heuristic way to generally
improve a target metric is to provide more relevant information to
the network. Previously we have exhausted all the information present
in a single invoice and so we will focus now on techniques related
to ``similarity''. Existing works on similarity are presented in
\prettyref{subsec:Inspiration} and our use and notion of similarity
is defined here in \prettyref{subsec:The-learning-framework}. In
short, we will present a similar annotated document as another input.
More details on differences from the previous work are described in
\prettyref{subsec:The-differences-to-prev}.

Since the idea of providing more information is fundamental even for
simpler templating techniques \cite{docsumo}, we need to stress that,
due to the nature of our dataset (which is available in anonymized
version at \cite{codeanddata}), our problem cannot be solved by using
templates. To prove this statement, a reasonable template-based baseline
will be presented (in \prettyref{subsec:Baselines}) and evaluated
(in \prettyref{sec:Experiments}).

The research question will focus on a ``similarity'' based mechanism
with various model implementations, and whether they can improve an
existing solution \cite{Ours}. The hypothesis is that we are able
to create at least one model that can significantly improve the results.
Moreover, since the presented mechanism is theoretically applicable
beyond the scope of document processing, this work can contribute
to a broader audience.

Ultimately we will present a model and its source code \cite{codeanddata}
that outperforms the previous state-of-art results. An anonymized
version of the dataset is also included as an open-source resource
and should be a notable contribution since its size is greater than
any other similar dataset known to date. 

\subsection{Related works\label{subsec:Related-works}}

This subsection focuses on research on previous works and approaches
in the relevant field of information extraction. The text in this
subsection is heavily based on the text from \cite{Ours}.

The plethora of methods that have been used historically for general
information extraction is hard to fully summarize or compare. Moreover,
it would not be fair to compare methods developed for and evaluated
on fundamentally different datasets.

However, we assessed that none of these methods is well-suited for
working with structured documents (like invoices), since they generally
do not have any fixed layout, language, caption set, delimiters, fonts...
For example, invoices vary in countries, companies and departments,
and change in time. In order to retrieve any information from a structured
document, you must understand it. Our criterion for considering a
method to compare against is that no human-controlled preprocessing
such as template specification or layout fixing is required because
we aim for a fully automated and general solution. Therefore we will
not be including any historical method as a baseline to compare against.

In recent works, a significant number does successfully use a graph
representation of a document \cite{8395204,Couasnon2014,KriegerInfExtract,RibaTableDetect,lohaniInvoice,liu2019graph}
and use graph neural networks. Also, the key idea close to the one-shot
principle in information extraction is used and examined for example
in \cite{10.1007/978-3-540-74141-1_28} and \cite{docsumo}. Both
works use notions of finding similar documents and reusing their gold-standards
(or already annotated target classes, if you will). The latter \cite{docsumo}
applies the principle in the form of template matching without the
need for any learnable parameters.

Our approach can also be called ``word classification'' approach
as written in \cite{Palm2019AttendCP}, a work where an end-to-end
architecture with a concept of memory is explored.

At this point, it is important to clarify the differences between
other works and our stream of research (meaning this work and previous
\cite{Ours}).

The most important difference comes from the dataset that is at our
disposal. The dataset explored here is far greater than the datasets
used elsewhere, and allows for exploring deeper models as opposed
to only using graph neural networks. Indeed in our previous paper,
we have proven that graph neural networks work in synergy with additional
convolution-over-sequence layers and even global self-attention. For
clarity, the roles of said layers are described in \prettyref{subsec:Common-architecture}.
Moreover, the dataset quality allowed us to discover (in the previous
paper) that information extraction and line-item table detection targets
do boost each other.

As the research is focused on deeper models, we will not be using
any of the other works as baselines and the commonly used graph neural
networks will be incorporated only as one layer amidst many, with
no special focus.

In the following pages, we will explore models that would be able
to benefit from access to a known similar document's page. We hope
that the model can exploit similarities between documents, even if
they do not have similar templates.

\subsection{Broader inspiration\label{subsec:Inspiration}}

A broader section on references is provided here since we are using
a great variety of layers in the exploration of deep network architectures.

\paragraph{One-shot learning and similarity\label{par:One-shot-learning-and}}

Presented in \cite{koch2015siamese,vinyals2016matching} is a model
design concept that aims to improve models on new data without retraining
of the network. 

Typically, a classification model is trained to recognize a specific
set of classes. In one-shot learning, we are usually able to correctly
identify classes by comparing them with already known data. Unlike
traditional multi-class classification, one-shot learning allows us
to attain better scores even with surprisingly low numbers of samples
\cite{fei2006one}. Sometimes it can work even for classes that are
not present in the training set \cite{vinyals2016matching}.

This concept can help in areas ranging from computer vision variants
– omniglot challenge \cite{lake2019omniglot} (also as strokes similarity
\cite{lake2011one}) to object detection \cite{xu2019recognition},
finding similar images \cite{Yim2018OneShotIS}, face detection \cite{facerecog},
autonomous vision \cite{grigorescu2018generative}, speech \cite{eloff2019multimodal}
and also the NLP area \cite{Yin2020MetalearningFF,kyle2018oneshot,dalvi2012websets}.

Among the methods that make one-shot learning able to work, the most
fundamental one utilizes the concept of similarity. For similarity
to work, we have two types of data – ``unknown'' and ``known''.
For the known data, its target values are known to the method and/or
to the model. To classify any unknown input, the usual practice is
to assign the same class to it as is the class of the most similar
known input.

Technically speaking, the architecture (typically) contains a “siamese”
part. In particular, both inputs (unknown and known) are passed to
the same network architecture with tied weights. We will draw inspiration
from this basic principle, and will leave other more advanced methods
of one-shot learning (for example, GANs \cite{mehrotra2017generative})
for further research.

Usually due to performance reasons the model is not asked to compare
new inputs to every other known input – only to a subset. Therefore,
a prior pruning technique needs to be incorporated – for example in
the form of the nearest neighbor search in embedding space, as is
done for example in the work \cite{ghosh2017r}. Another option would
be to incorporate a memory concept \cite{cai2018memory} (even in
the form of neural Turing machines \cite{santoro2016one}).

The loss used for similarity learning is called triplet loss because
it is applied on a triplet of classes ($R$ reference, $P$ positive,
$N$ negative) for each data-point:

\begin{align*}
L(R,P,N)=\min( & \left\Vert f(A)-f(P)\right\Vert ^{2}-\\
 & \left\Vert f(A)-f(N)\right\Vert ^{2}+\alpha,0)
\end{align*}
Where $\alpha$ is a margin between positive and negative classes
and $f$ is the model function mapping inputs to embedding space (with
euclidean norm).

Generally speaking, one-shot learning can be classified as a meta-learning
technique. For more on meta-learning, we suggest a recent study, like
\cite{peng2020comprehensive} (or just a compiled bibliography online
at \cite{MetaLearningPapers}). Taking the concept one step further
yields a concept called ``zero-shot learning'' \cite{paul2019semantically,felix2019multi,mishra2018generative}.

\paragraph{Other sources of inspiration\label{par:Other-sources-of}}

It is now beneficial to mention other sources of inspiration that
are also meaningfully close to one-shot learning. Since we ask ``what
labels are similar in the new data'', a ``query answer'' approach
should be considered. Recently, the attention principle (namely the
transformer architecture) successfully helped to pave the way in language
models \cite{radford2019language}. It is not uncommon to use attention
in one-shot approaches \cite{wang2017multi}  and also query answer
problems in various problems domains \cite{nie2017attention,Galassi2020AttentionIN,Yang2016StackedAN}. 

The mentioned task of similarity can also be approached as pairwise
classification, or even dissimilarity \cite{pairwiseDissim10.1007/978-3-540-89639-5_3}.

\section{Methodology\label{sec:Methodology}}

We want to explore the added effect of ``similarity'' while keeping
everything as close to the previous setting as possible to make sure
no other effect intervenes. 

To not require the reader’s knowledge of the previous work, we need
to establish common grounds at the beginning of this section (in \prettyref{subsec:Overview}
and \prettyref{subsec:Common-architecture}). Therefore the following
description mirrors the description given in previous work (in 3.3
and 3.4 of \cite{Ours}). Note that previous work was not using any
means of ``similarity'' or ``nearest pages'' and so they are introduced
first in \ref{subsec:The-learning-framework} (If the reader happens
to be familiar with the content, they can continue reading there.)

\subsection{Overview\label{subsec:Overview}}

The main unit of our scope is every individual word on every individual
page of each document. Note that other works (such as \cite{liu2019graph})
use the notion of “text segments’ instead of “words”. For the scope
of this work, we define a “word” as a text segment that is separated
from the rest of the text by (at least) a white-space, and we will
not consider any other text segmentation. 

\paragraph{Inputs and outputs\label{par:Inputs-and-outputs}}

Conceptually, the whole page of a document is considered to be the
input to the whole system. Specifically, the inputs are the document's
rendered image, the words on the page and the bounding boxes of the
words. As PDF files are considered, any possible library for reading
PDF files can be used for reading the files and getting the inputs.
Note, that also by using any standardized OCR technique, the method
could theoretically be applied to scanned images too (measuring the
effect of OCR errors on the extraction quality is not done here).

These inputs then undergo a feature engineering described in \prettyref{subsec:Basic-description-of}
and become inputs for a neural network model.

Each word, together with its positional information (“word-box,” in
short) is to be classified into zero, one, or more target classes
as the output. We are dealing with a multi-label problem with 35 possible
classes in total. The classes include the ``total amount'', tax
information, banking information, issuer, and recipient information
(the full set being defined in the code \cite{codeanddata}). To obtain
a ground-truth, the classes were manually annotated by expert human
annotators. Interestingly, a human has roughly $3\,\%$ error, which
was eliminated by a second annotation round.

\paragraph{The dataset and the metric\label{par:The-dataset}}

Overall, we have a dataset with $25\,071$ documents as PDF files
totalling $35\,880$ pages. The documents are of various vendors,
layouts, and languages, and are split into a training, validation,
and test set at random ($80\,\%$ / $10\,\%$ / $10\,\%$). 

We will observe and report the scores of a testing set. A validation
set is used for model selection and early stopping. The metric used
is computed first by computing all the $F_{1}$ scores of all the
classes and aggregated by micro metric principle (more can be found
for example in \cite{narasimhan2016optimizing}) over all the word-boxes,
over all the pages.

The metric choice is inspired by the work \cite{gobel2013icdar} where
a content-oriented metric was defined on a character level. In our
setting the smallest unit is a word-box. The choice of $F_{1}$ score
is based on the observation that the counts of positive samples are
outnumbered by the negative samples - in total, the dataset contains
$1.2\,\%$ positive classes.

\subsection{Shared architecture parts\label{subsec:Common-architecture}}

\begin{figure}
\includegraphics[width=1\columnwidth]{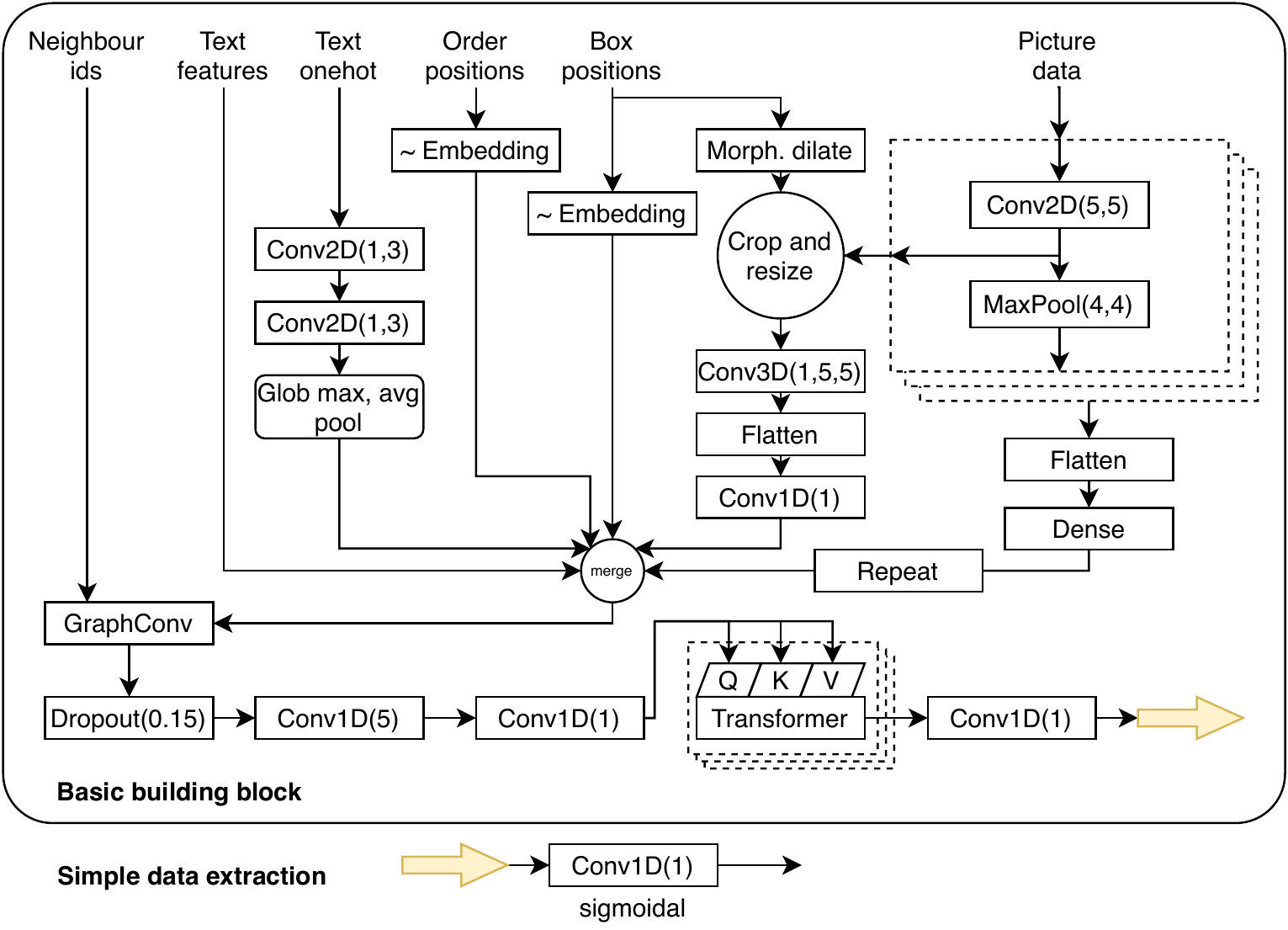}

\caption{\label{fig:Simple-data-extraction}Simple data extraction model. Formally
the whole model consists of two parts: a Basic building block and
a final classification layer. The whole model is formally split into
two parts, as the Basic building block will be used (as a siamese
network) in other models. By removing the final classification layer,
we hope to get the best feature representation for each word-box.}
\end{figure}

We will call the architecture from the previous work a ``Simple data
extraction model'' and it will be one of the baselines in this work.
The architecture of the model is the same as in the previous work
and is varied only by a minor manual parameter tuning. A notable part
of this model (called ``Basic building block'') will be used here
in all the new models (defined in section \ref{subsec:Model-architectures}).
Both the Simple data extraction model and Basic building block are
depicted in \prettyref{fig:Simple-data-extraction}. 

Since the overall task's goal and the whole ``Basic building block''
architecture are shared across all models, by describing the ``Simple
data extraction model'', we define and describe all of the shared
and inherited parts – notably the input and output requirements.
We are using full geometrical, visual, and textual information as
the input, and the model outputs a multi-class classification for
each word-box. We will proceed by describing the processing of the
inputs in detail.

\subsubsection{Detailed feature engineering of the inputs\label{subsec:Basic-description-of}}

We will operate based on the principle of reflecting the structure
of the data in the model's architecture, as Machine learning algorithms
tend to perform better that way.

The structured information at the input will be an ordered sequence
of all the word-boxes present on a page. Note that the number of word-boxes
per page can vary.

\begin{figure}
\includegraphics[width=1\columnwidth]{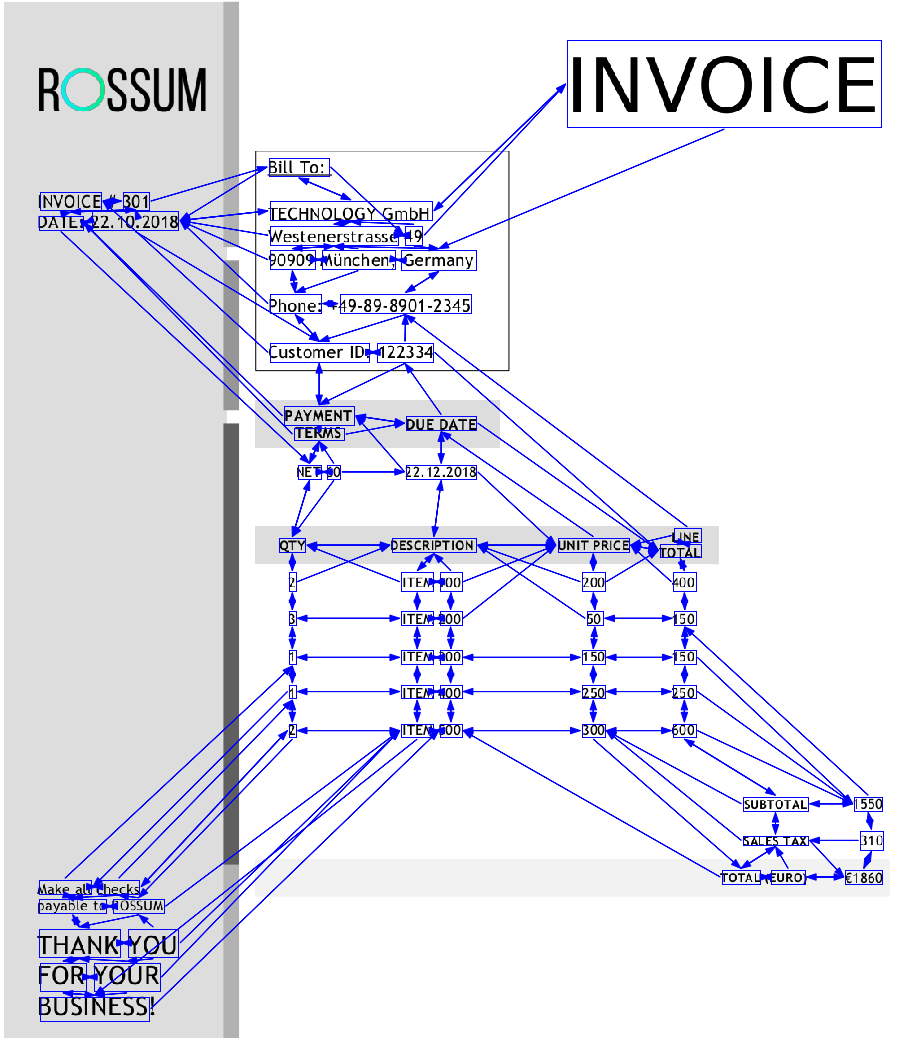}\caption{\label{fig:Sample-invoice-edges}Sample invoice with edges defining
neighborhood word-boxes. Only the closest neighbor is connected for
each word-box. (This invoice was artificially created for presentation
and does not represent the invoices in the dataset.)}
\end{figure}

The features of each word-box are:
\begin{itemize}
\item Geometrical:
\begin{itemize}
\item Using a geometrical algorithm we can construct a neighborhood graph
over the boxes, which can then be used by a graph CNN (see \prettyref{par:Graph-convolution-mechanism}).
\\
Neighbors are generated for each word-box ($W$) as follows – every
other box is formally assigned to an edge of $W$ that has it in its
field of view (being fair $90\text{°}$), then the closest (center
to center Euclidean distance) $n$ neighbors are chosen for that edge.
For example with $n=1$ see \prettyref{fig:Sample-invoice-edges}.
The relation does not need to be symmetrical, but when a higher number
of closest neighbors is used, the sets would have a bigger overlap.
\item We can define a ’reading order of word-boxes’. In particular, based
on the idea that if two boxes do overlap in a projection to $y$ axis
by more than a given threshold, set to $50\,\%$ in the experiments,
they should be regarded as being in the same line for a human reader.
This not only defines the order in which boxes will be given as a
sequence to the network, but also assigns a line number and order-in-line
number to each box. To get more information, we can run this algorithm
again on a $90\text{°}$ rotated version of the document. Note that
the exact ordering/reading direction (left to right and top to bottom
or vice versa) should not matter in the neural network design, thus
giving us the freedom to process any language.
\item Each box has 4 normalized coordinates (left, top, right, bottom) that
should be presented to the network.
\end{itemize}
\item Textual:
\begin{itemize}
\item Each word can be presented using any fixed-size representation. In
this case, we will use tailored features common in other NLP tasks
(e.g. authorship attribution \cite{DBLP:journals/corr/abs-1208-3001},
named entity recognition \cite{nadeau2007survey}, and sentiment analysis
\cite{Abbasi:2008:SAM:1361684.1361685}). The features per word-box
are the counts of all characters, the counts of the first two and
last two characters, length of a word, number of uppercase and lowercase
letters, number of text characters and number of digits. Finally,
another feature is engineered to tell if the word is a number or amount.
The new feature is produced by scaling and min/maxing the amount by
different ranges. (If the word is not a number, this feature is set
to zero.) We chose all these features because invoices usually include
a large number of entities, ids, and numbers that the network needs
to be able to use.
\item Trainable word features are employed as well, using convolutional
architecture over a sequence of one-hot encoded, deaccented, lowercase
characters (only alphabet, numeric characters and special characters
`` ,.-+:/\%?\$£€\#()\&'{}'', all others are discarded). We expect
these trainable features to learn the representations of common words
that are not named entities.
\end{itemize}
\item Image features:
\begin{itemize}
\item Each word-box has its corresponding crop in the original PDF file,
where the word is rendered using some font settings and also background.
This could be crucial to a header or heading detection, if it contains
lines, for example, or different background color or gradient. So
for each word-box, the network receives a crop from the original image,
offset outwards to be bigger than the text area to also see the surroundings.
\end{itemize}
\end{itemize}
Each presented feature can be augmented, we have decided to do a random
one percent perturbation on coordinates and textual features.

\subsubsection{Simple data extraction model details}

It is now convenient to summarize the document's features described
in the previous section, since we will now explain how they are processed
by the model (as Figure \vref{fig:Simple-data-extraction} shows).
In total we have $5$ inputs the neural networks will use:
\begin{itemize}
\item Down-sampled picture of the whole document ($620\times877$), gray-scaled.
\item Features of all word-boxes (as defined in the previous section), including
their coordinates.
\item Text as first 40 one-hot encoded characters per each word-box.
\item Neighbor ids – lookup indexes that define the neighboring word-boxes
on each side of the word-box.
\item And finally the integer positions of each word-box defined by the
geometrical ordering.
\end{itemize}
In the Simple data extraction model, the positions are embedded by
positional embeddings (as defined in \cite{2017arXiv170603762V,2018arXiv180703247L}.
Embedding size equal to $4$ dimensions for $sin$ and $cos$, with
divisor constant being $10000$ is used. The embedded positions are
then concatenated with other word-box features.

The picture input is reduced by a classical stacked convolution and
max-pooling approach. The word-box coordinates (left, top, right,
bottom) are not only used as a feature, but also to crop the inner
representation of the picture input (see ``Morphological dilation''
in \prettyref{fig:Simple-data-extraction}). Finally, we have decided
to give the model a grasp of the image as a whole and to supply a
connection to the said inner representation – flattened and then processed
to $32$ float features.

Before attention, dense, or graph convolution layers are used, all
the features are simply concatenated. To accompany this description,
equations and network definitions are given in\cite{codeanddata}.

Like the previous work shows, all three means of assessing relations
between word-boxes are used:
\begin{itemize}
\item Graph convolution (also denoted as ``GCN'') over the geometrical
neighbors of word-boxes is employed to exploit any form of local context
(detailed explanation is given at the end of this section in \nameref{par:Graph-convolution-mechanism}).
\item A 1D convolution layer (called \emph{convolution over sequence} in
previous work) over word-boxes ordered by the reading-order allows
the network to follow any natural text flow. Implementation-wise all
the word-boxes are ordered in the second dimension at the input (all
dimensions being {[}batch, ordering, feature space{]}).
\item The attention transformer module (from \cite{2017arXiv170603762V})
allows the network to relate word-boxes across the page. Our attention
transformer unit does not use causality, nor query masking.
\end{itemize}

After these layers are applied, the Basic building block definition
ends with each word-box embedded to a feature space of a specified
dimension (being 640 unless said otherwise in a specific experiment).
The following layer, for the ``Simple data extraction model'', is
a sigmoidal layer with binary cross-entropy as the loss function.
This is a standard setting, since the output of this model is meant
to solve a multi-class multi-label problem.

To note implementation detail, batched data fed to the model are padded
by zeros per batch (with zero sample weights). Class weights in the
multi-task classification problem were chosen (and manually tuned)
based on positive class occurrences. 

\paragraph{Graph convolution mechanism details\label{par:Graph-convolution-mechanism}}

The word-box graph (with word-boxes as nodes and neighborhood relation
as edges, as depicted in \prettyref{fig:Sample-invoice-edges}) has
a regularity that allows us to simplify the graph convolution. First,
there does exist a small upper bound on the number of edges for each
node, and second, we do not desire to use any edge classification
or specific edge features unlike other works (like \cite{RibaTableDetect}).
Therefore we use a simpler implementation than the general form graph
convolutions (as in \cite{2016arXiv160505273N,2017arXiv170802218J}).

In detail, the implementation uses a generic simplicity present in
convolutions at the cost of an additional input. Even a classical
convolutional layer over regular picture data can be represented by
two basic operations. First, a gather operation (using tf.gather\_nd
function from \cite{tensorflow2015-whitepaper}) that prepares the
data to a regular array (matrix of size number of data points times
the number of data points in one convolutional operation). The second
operation would then be a time-distributed dense (or we can call it
conv1d as these two names are the same for 1d sequential data) layer
that simulates the weights of such convolution.

The gather operation needs additional input for each point (pixel
or graph node) that specifies the integer indexes of its neighbors
(and the node itself). These integer indexes are constructed exactly
as stated in \ref{subsec:Basic-description-of}.

\subsubsection{The differences to previous setting\label{subsec:The-differences-to-prev}}

At this point, we have hopefully cited enough from the previous research
to make the basics clear. Just as we have noted the differences to
existing research in \prettyref{subsec:Related-works}, it is also
important to note some detailed differences to previous work.

\paragraph{The novelty of this work w.r.t the previous setting}

The previous work \cite{Ours} did not use any nearest neighbor search,
nor models that would utilize any notions of similarity or more than
one input page at once. In short, the previous work just lays the
fundamental principles of the data, task, metric and introduces the
Basic building block (with ablation analysis) as explained up to this
point in this section \prettyref{subsec:Common-architecture}. Everything
else that follows is new.

\paragraph{Details changed from the previous setting}

Unlike in the previous setting, we will not be classifying the line-item
tabular structures, but only extracting (above mentioned) information
from the page. This should also demonstrate that the model, even though
previously optimized on line-item table detection, is versatile enough.
Hence we will allow ourselves only minor tweaks in the model's architecture
(results of the modifications depicted in the diagram \ref{fig:Simple-data-extraction}).

Previously, two datasets were used for training and validation - ``small''
(published) and ``big'' (previously unpublished). The models were
tuned on the small dataset and the big one was used only in two experiments
to validate that the model scales. In this work, the same big dataset
is used, it's previous validation set is split into a new validation
and a new test set to make the test set bigger and properly address
generalization (see the next part in – \ref{subsec:The-differences-to}).

Multiple baselines are employed to prove that the new test set contains
documents with layers that are different enough (previous work's test
set was small and manually selected).

\subsubsection{The differences to one-shot learning\label{subsec:The-differences-to}}

As stated in the introduction, we want to boost the performance for
existing target classes by giving the network access to ``known''
data (documents) in ways similar to one-shot learning. The main difference
is that we will be utilizing experiments and architectures that would
include a fixed selection of classes and/or class information (from
the nearest page). It is important to clarify this detail since usually
in one-shot learning, no classes are explicitly present in the model
as the aim is to generalize to those classes. Our aim, by contrast,
is to generalize to different and unseen documents with different
layouts (instead of classes) that still feature those word-box classes.

\subsection{The learning framework\label{subsec:The-learning-framework}}

Let’s now look at the big picture of the proposed method. 

The easiest step to boost predictions of an unknown page is to add
one more page that is similar and includes word-box classes (annotation)
already known to the system to use that annotation information in
a trained model.

Overall the method would work as follows:
\begin{itemize}
\item The system needs to keep a notion of already ``known'' documents
in a reasonably sized set. We call them ``known'', as their classes/annotations
should be ready to use.
\item When a ``new'' or ``unknown'' page is presented to the system,
search for the most similar page (given any reasonable algorithm)
from the ``known'' pages. 
\item Allow the model to use all the information from both pages (and ``learn
from similarity'') to make the prediction.
\end{itemize}
The system can then even present the predictions to a human to verify
them and then add the page to the existing database of known pages,
but we will not be exploring the database size effects here.

Before predicting, the incorporated model should be trained on pairs
of pages simulating this behavior.

In this process, there are more points to be examined, but we believe
that the most interesting research question is the following:

Holding all other factors fixed (meaning the train\slash test\slash validation
split, evaluation metrics, data format, and method for searching for
a similar page), what approach and what neural network architecture
is able to raise the test score the most?

We believe that this is the right question to ask since all other
factors have usually a known effect on the result if best practices
are followed. As an example, we ask the reader to recall that bigger
datasets typically bring better scores, having more ``nearest neighbors''
typically has a boosting effect similar to ensembling, and so on.

Also from a practical point of view, only two pages can fit into a
single GPU memory with all the features described before.

As stated in the introduction, we will draw inspiration from the one-shot
learning framework. For predicting an unknown page, we will define
a way to search for one ``nearest'' known page and allow the model
access to its annotations as known target classes. Note that not all
explored models will use the nearest known page – in addition to the
Simple data extraction model, we will consider some baselines that
do not require the nearest page to verify the assumptions.

\paragraph{Nearest neighbor definition\label{par:Nearest-neighbor-definition}}

For one-shot learning to work on a new and unknown page (sometimes
denoted ``reference''), the system always needs to have a known
(also denoted ``similar'' or ``nearest'') document with known
annotations at its disposal. Since focusing on that task properly
would be material for another paper, we have used the nearest neighbor
search in the space of the page`s embeddings to select only one closest
page of a different invoice document.

The embeddings were created through a process similar to a standard
one, which is described in \cite{burkov2020machine}. We have used
a different model (older and proprietary) that was trained to extract
information from a page. To change the classification model into an
embedding model, we have removed its latest layer and added a simple
pooling layer. This modified the model to output $4850$ float features
based only on image input. These features were then assigned to each
page as its embedding. 

We then manually verified that the system would group similar, or
at least partially similar, pages near each other in the embedded
space.

These embeddings are held fixed during training and inference and
computed only once in advance.

\paragraph{Constraints of nearest neighbor search\label{par:Constraints-of-nearest}}

We want the trained model to behave as close to the real world as
it can, so the nearest page search process needs to be constrained.
Each document's page can select the nearest annotated page only from
the previous documents in a given order. As in a real service, we
can only see the arrived and processed documents.

Also, we want the method to be robust, so before each epoch, the order
of all pages is shuffled and only the previous pages (in the said
order) from a different document are allowed to be selected. 

This holds for all sets (training, validation and test) separately.
To verify the consistency of this strategy, some experiments will
be tweaked by the following variations:
\begin{itemize}
\item Allowed to additionally use the training set as a data source for
the ``nearest annotated'' input. We expect the performance to rise.
\item Made ``blind'' by selecting a random document's page as the nearest
known input. We expect the performance to fall.
\end{itemize}

\subsubsection{Baselines\label{subsec:Baselines}}

To challenge our approach from all possible viewpoints, we will consider
multiple baselines:
\begin{enumerate}
\item To use only the ``Simple data extraction model'' (described in section
\prettyref{subsec:Common-architecture} and \prettyref{fig:Simple-data-extraction})
without any access to the nearest known page.
\item ``Copypaste'' baseline. This model will only take the target classes
from the nearest page's word-boxes and overlay them over the new page's
word-boxes (where possible). We expect a low score since the documents
in the dataset are different, and this operation will not copy anything
from any nearest page's word-box that does not intersect with a new
page's word-box. This approach uses no trainable weights and happens
to be the simplest example of a templated approach that does not have
hard-coded classes.
\item ``Oracle'' baseline. This model will always correctly predict all
classes that are present in the nearest page. We use this model to
measure the quality of the nearest-page embeddings to gain additional
insight into the dataset's properties. The metric used for this model
is not $F_{1}$, but a percentage of all word-boxes that can be classified
correctly. The score is expected to be only moderately good, as the
embeddings are created in a rather unsupervised manner (regarding
their usage). We want to explore a different influence than already
existing works aimed at finding the best helping pages \cite{docsumo}
do explore. Ultimately we want to present a model that can work even
if the quality of the embeddings is just moderate.
\item Fully linear model with access to concatenated features from both
new and known pages. Does not feature picture data.
\end{enumerate}
The choice of baselines (and ablations later in experiments) is as
such to verify and demonstrate multiple claims:
\begin{itemize}
\item It can beat the previous results.
\item The documents are different enough.
\item pJust similarity search alone is not enough, even if the embeddings
would have better than moderate quality with regard to the similarity.
\item To justify the complexity of models presented in the following section,
\ref{subsec:Model-architectures}.
\end{itemize}
To elaborate on the last point – the Copypaste baseline represents
a reasonable basic counterpart for ``Triplet loss'' and ``Pairwise
classification''. The fully linear model represents the simplest
counterpart for the ``query answer'' approach that also has all
the classes hardcoded (these architectures and terms are defined in
\ref{subsec:Model-architectures}).

All baselines and all models presented here will have the same desired
output – they will provide the multi-class classification for each
word-box.

\subsection{Model architectures\label{subsec:Model-architectures}}

We have described the Basic information extraction block that aims
to output the best trained latent features for each word-box. All
the model architectures will incorporate this block used as a siamese
network for the inputs of both unknown and known pages. Every single
one of the architectures is trained as a whole, no pre-training or
transfer learning takes place, and every model is always implemented
as a single computation graph in tensorflow. 

We will explore multiple different architectural designs that will
be predicting the targets (at their outputs) by using the closest
nearest page from already annotated documents. 
\begin{enumerate}
\item ``Triplet Loss architecture'' – using siamese networks ``canonically''
with triplet loss.
\item ``Pairwise classification'' – using a trainable classifier pairwise
over all combinations of word-box features from reference and nearest
page. 
\item ``Query answer architecture'' (or ``QA'' for short) – using the
attention transformer as an answering machine to a question of ``which
word-box class is the most similar''.
\end{enumerate}
There is a slight distinction between the first two and the last architecture.
In Query answer architecture the class is a direct prediction of the
network for each word-box. In Triplet loss and Pairwise classification,
the models predict (for each unknown word-box) all the similarities
to all the word-boxes from the known page. All the similarity values
then vote in an ensemble way for the target class for the word-box.

Since the embeddings used to search for the nearest page are not ideal,
there might be some classes that the models would not be able to predict.
To assess these methods fairly, we will scale the metrics used to
measure the success by the performance of the corresponding Oracle
baseline (defined in \prettyref{subsec:Baselines}). Or put differently,
we would not be counting errors that the model would not be able to
predict correctly due to some classes being absent from the nearest
page. This reflects the aim to explore the effects of the models that
can operate with the nearest page.

In reality, if these (1. and 2.) methods would be proved most efficient,
the hyperparameters, such as the quality of the embeddings (or the
number of the nearest pages), would need to be addressed to overcome
the score of the previous results. The perfect performance of the
scaled metric means that the extraction is only as good as the Oracle
baseline.

In the experimental results section below (\ref{sec:Experiments}),
we will include a test of models 1. and 2. that would make the models
predict a different and possibly easier target. Instead of ``do these
two word-boxes have the same target class'', the easier testing target
would be ``do these two word-boxes have the same length of text inside''.
This test is meant to show that the method is well-grounded and usable
for any other reasonable target definition. 

On the other hand, Query answer architecture has the classes hard-coded
in the design, which means it can predict a class not present in the
nearest page. Therefore no metric scaling takes place in the evaluation
of the QA model.

\subsubsection{Triplet loss architecture\label{subsec:Triplet-loss-architecture}}

\begin{figure}
\includegraphics[width=1\columnwidth]{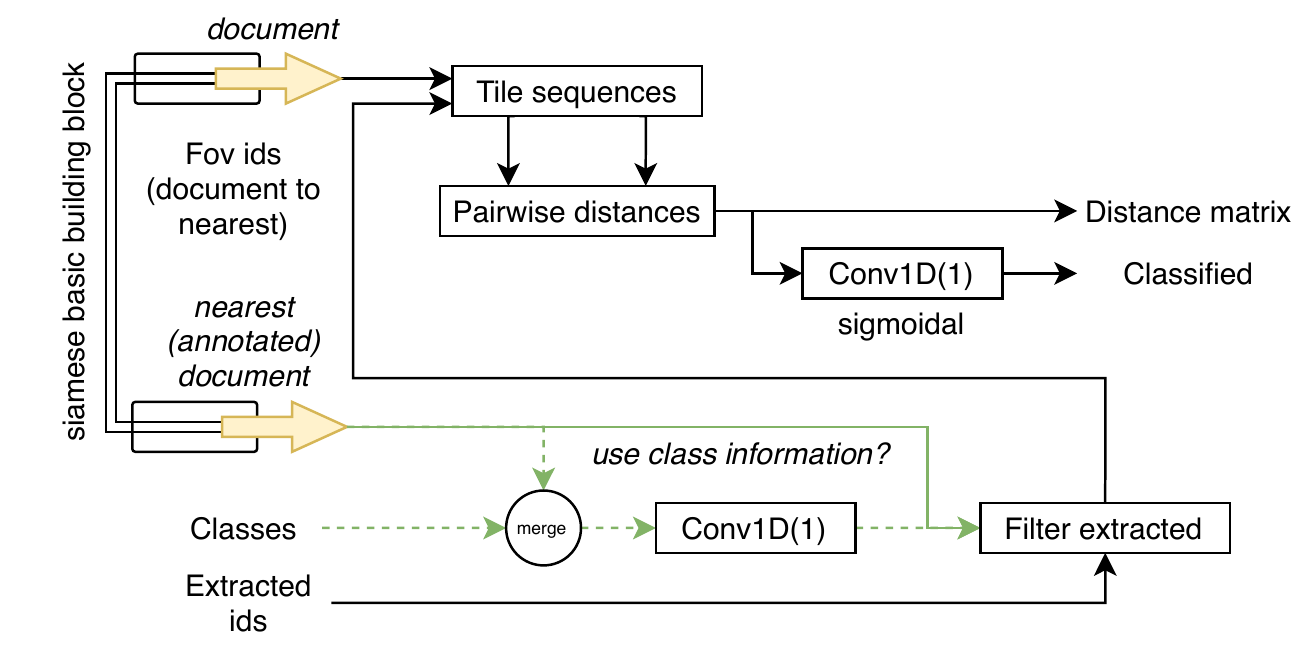}

\caption{\label{fig:triplet}Triplet loss architecture. If we want to add class
information from the nearest page, the green dashed version is used.}
\end{figure}

Since our data-point is a word-box, adhering strictly to using triplets
of word-boxes for triplet loss would force us to execute the model
for each word-box pair once. To not impair the performance (as there
can be as many as 300 word-boxes per page) and/or lose in-page dependencies,
the proposed architecture (see \prettyref{fig:triplet}) features
a mechanism of tiling and filtering to pass all combinations of word-boxes
at once.

The filtering mechanism filters only the annotated word-boxes from
the nearest page. It eliminates most of the unused information and,
in doing so, saves memory and computation time. The tiling mechanism
takes two sequences – first, the sequence of reference page word-boxes
and second the sequence of nearest page filtered word-boxes; and produces
a bipartite matrix. The model is then able to compute pairwise distances
between the same and different classes. These distances are then used
for triplet loss computation (see mathematical definition in the section
below).

Additionally, we can include a single classification layer to be automatically
calibrated on the distances, which adds a binary cross-entropy term
to the loss. Because there are more word-boxes on each page, the loss
is averaged over all the word-boxes.

We rely on the (manually verified) fact that during training each
page has more than 1 class annotated. And because of that, there are
always positive and negative samples present, as there should be in
the triplet loss.

There are three possible modifications to explore:
\begin{itemize}
\item Adding annotated class information to the nearest page's features.
\item Using a ``loss-less triplet loss''. A loss similar to the triplet
loss but without the min-max functions (see definition below).
\item Modifying the distance and/or loss computations by the means of constants
or by using cosine similarity instead of euclidean space. 
\end{itemize}

\subsubsection{Triplet-loss inspired losses}

The purpose of this model is to use the triplet loss most straightforwardly
in our setting. The only mathematically interesting description to
be given here is the triplet loss and ``loss-less triplet loss''
defined over word-boxes since all trainable layers in this model (and
binary cross-entropy loss) are defined in referenced works. 

In traditional triplet loss, we need positive, negative and reference
samples. Since we need to account for a whole page full of word-boxes,
we must compute all combinations at once.

We denote the quantity $\text{truth\_similar}(i,j)$ to indicate if
the word-boxes $i,j$ ($i$-th being from the unknown page, $j$-th
being in the nearest page) do share the same ground truth class ($1.0$
= yes, $0.0$ otherwise). Next we define $\text{pred\_dist}(i,j)$
as the predicted distances between the feature spaces of the word-boxes
by the model. Then we can calculate two loss variants (``triplet\_like''
and ``loss-less'') inspired by triplet loss as follows:

\begin{align*}
\text{pos\_dist}_{i,j}= & \text{truth\_similar}(i,j)\cdot\text{pred\_dist}(i,j)\\
\text{neg\_dist}_{i,j}= & (1.0-\text{truth\_similar}(i,j))\cdot\text{pred\_dist}(i,j)\\
\text{triplet\_like}= & \text{max}(0,\,\alpha+\text{max}(\text{pos\_dist}_{i,j})\\
 & +\text{min}(-\text{neg\_dist}_{i,j}))\\
\text{lossless}= & \sum_{i,j}\text{pos\_dist}_{i,j}-\sum_{i,j}\text{neg\_dist}_{i,j}
\end{align*}

Where pos\_dist and neg\_dist are just helper variables to see the
similarity with the original triplet loss, and $\alpha$ is a parameter
of the same meaning as in the original triplet loss. The two new losses
represent two different approaches used in the reduction from a matrix
to a single number. We can either take the largest positive and negative
values and use them in the triplet loss equation, or we can sum all
the positive and negative terms. The real difference is how the gradients
are propagated, variants with min/max always propagate fewer gradients
than the former per gradient-update step in the training phase. All
the losses can be used at once with a simple summation.

The name ``loss-less'' comes from the idea described in \cite{LosslessTriplet}
(and, to date, is not to be found in any scientific work other than
this online article). 

To wrap up this paragraph – we present different options for the loss
terms. Since we focus on different architectures and not on hyperparameters,
we omit from this description the specific constants used to sum the
terms up. In the experiments section (\ref{sec:Experiments}) the
results are the best that we were able to achieve by manual hyperparameter
tuning. The results of the tuning and various options are clearly
defined in the accompanying code \cite{codeanddata} together with
all the specific hyperparameters.

\subsubsection{Pairwise classification\label{subsec:Pairwise-classification}}

\begin{figure}
\includegraphics[width=1\columnwidth]{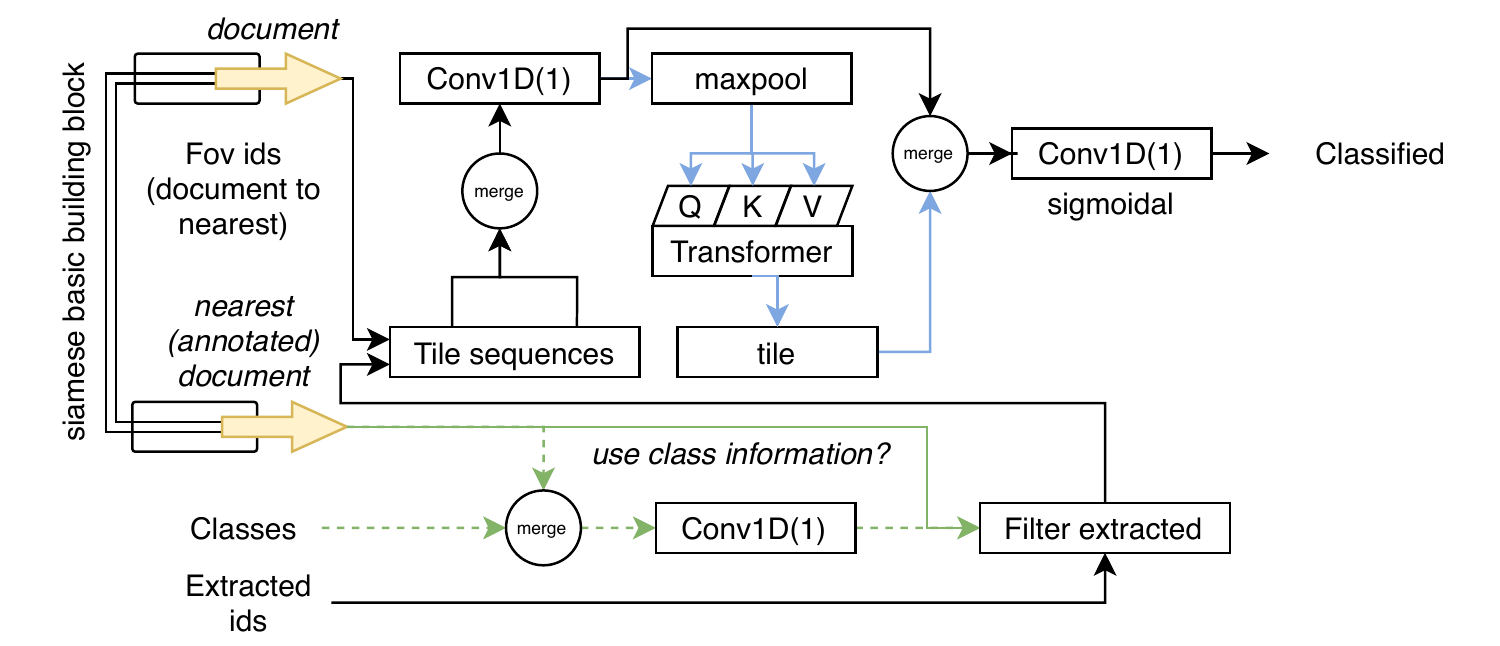}

\caption{\label{fig:pairwise}Pairwise classification architecture with an
optional refinement module.}
\end{figure}

This architecture (see \prettyref{fig:pairwise}) uses the same tiling
and filtering mechanism as we have described before in \ref{subsec:Triplet-loss-architecture}.
But instead of projecting the data points into a specific feature
space to compute distances, they are simply ``classified'' by using
a traditional approach of sigmoidal activation function and binary
cross-entropy loss.

Like in the previous model, we have the option of adding annotated
class information to the nearest page's features. We have also explored
various sample weights options and an optional ``global refinement''
section. The optional refinement pools information from each word-box,
uses a global transformer and propagates the information back to each
reference word-box – nearest word-box pair to be classified with the
refinement information available.

\subsubsection{Query answer architecture\label{subsec:Query-answer-architecture}}

\begin{figure}
\includegraphics[width=1\columnwidth]{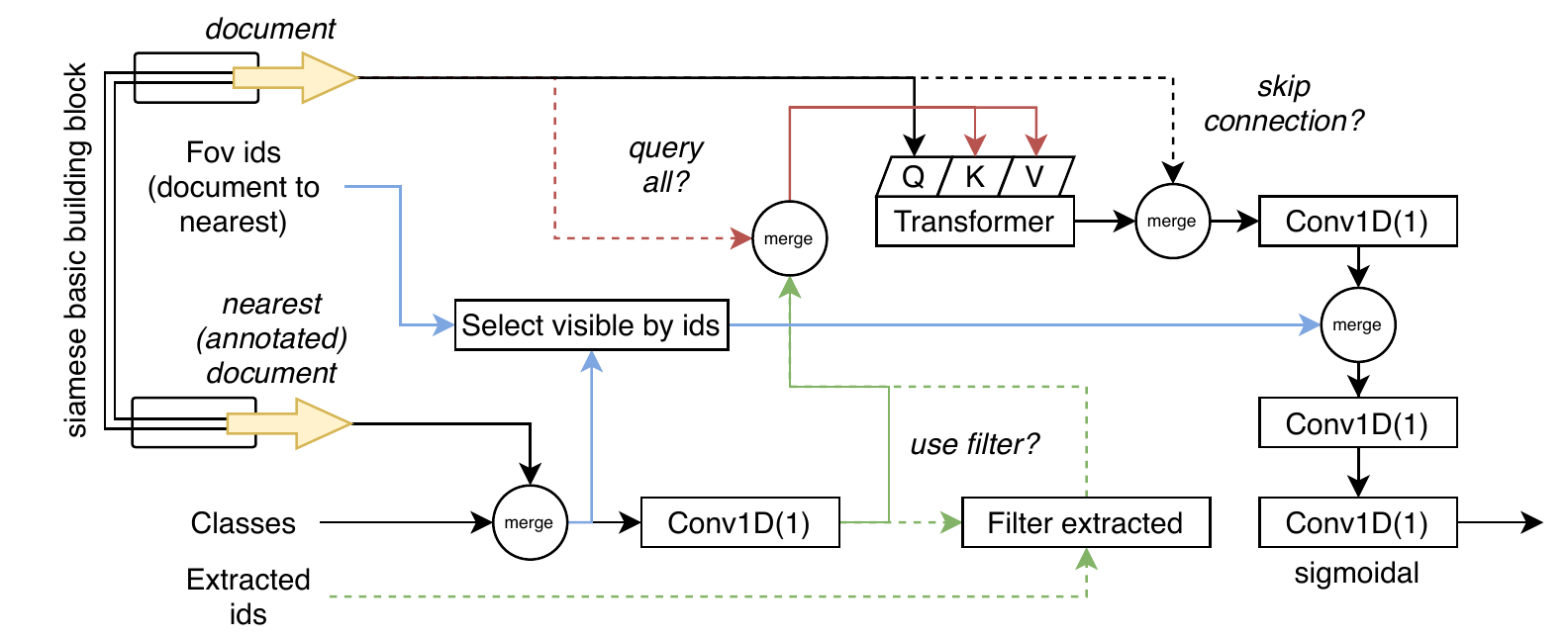}

\caption{\label{fig:qaa}Query answer architecture}
\end{figure}

In the heart of this architecture (see \prettyref{fig:qaa}) lies
the fact that the Transformer module with three inputs can be used
as a query-answer machine.

More variants could be explored here:
\begin{itemize}
\item ``Query all'': does it help if the Transformer can query not only
the nearest page's word-boxes, but also those of the new page itself?
\item ``Skip connection'': would a skip connection to the base information
extraction block improve the performance?
\item ``Filter'': should it filter only annotated word-boxes from the
nearest page? (As in the two previous approaches.)
\item ``Field of view'': would adding a field of view information flow
from the new page's word-boxes to the nearest page make a difference?
\end{itemize}
Technically a field of view is realized by providing indexes, which
word-boxes would be close to each other by geometrically projecting
each word-box from the reference page to the annotated page and selecting
a fixed number of euclidean-closest word-boxes. The limits for the
distances were chosen based on average distances between word-boxes
of the same class on different documents. The loss used for this model
is classical binary cross-entropy.

The main idea of this architecture is a query answer mechanism and
so it can be applied in any different setting with siamese networks. 

\section{Experiments and results\label{sec:Experiments}}

In this section we will cover the experimental results for each group
of experiments. Adam optimizer was used together with an early stopping
parameter of $20$ epochs (to maximally $200$ epochs). The average
time was $40$ minutes per epoch on a single GPU. The baseline needed
only $10$ minutes per epoch (since it did not need any ``nearest''
page mechanism). The model selected in each experimental run was always
the one that performed best on the validation set in terms of loss.

The Basic building blocks present in every architecture were usually
set to produce feature space of dimensionality $640$ (unless noted
otherwise in the tables as ``feature space $n$'').

Additionally, experiments on the anonymized dataset are performed
on the best architecture and the baseline model. The anonymized dataset
does not include picture information, and textual information there
is replaced by letters ``a'' of similar length. Moreover, some features
in some documents are randomly adjusted in various ways, so there
is no reliable way of mapping the anonymized documents to reality. 

Some experiments with architecture variations are included to show
how the model's variance affects the score – for that reason, we have
slightly varied the number of the transformer layers – (``1x attention
layer'' marks single layer, ``2x attention layer'' marks two consecutive
layers being used), as that is the single most complex layer present.

\subsection{Baselines results}

We are reporting some variations of architecture parameters for the
Simple data extraction model (introduced in section \prettyref{subsec:Common-architecture})
in the \prettyref{tab:Ex-Res-SImple-Model}. The goal is to show how
much the basic model is sensitive to various changes and to give the
baseline some chance to be tuned for extracting the classes. 

 The results could be interpreted as the model reaching its maximal
reasonable complexity at 1 transformer layer and smaller feature space.
As we will see, this does not apply to the siamese settings as the
gradients propagate differently when parts of the architecture have
tied weights.

To beat the previous state of the art results, we need to improve
$F_{1}$ score over $0.8465$, which is the best score for the Simple
data extraction model.

\begin{table}[t]
\caption{\label{tab:Ex-Res-SImple-Model}Simple data extraction model experimental
results. }

\begin{tabular*}{1\columnwidth}{@{\extracolsep{\fill}}|>{\raggedright}p{0.8\columnwidth}|>{\centering}p{0.1\columnwidth}|}
\hline 
\multirow{1}{0.8\columnwidth}{Previous state of the art, re-tuned\\
(and possible notable tweaks, see section \ref{sec:Experiments})} & Test micro\\
$F_{1}$ score\tabularnewline
\hline 
\hline 
2x attention layer, feature space 640 & 0.6220\tabularnewline
\hline 
1x attention layer, feature space 640  & 0.8081\tabularnewline
\hline 
1x attention layer, feature space 64  & \textbf{0.8465}\tabularnewline
\hline 
1x attention layer, f. space 64, fully anonymized  & 0.6128\tabularnewline
\hline 
1x attention layer, f. space 64, only text features  & 0.7505\tabularnewline
\hline 
\end{tabular*}
\end{table}

\paragraph{Copypaste baselines}

\prettyref{tab:Copypaste-baseline-results} shows the fairly low score
of those simple baselines. Such a low score illustrates the complexity
of the task and variability in the dataset. Simply put, it is not
enough to just overlay a different similar known page over the unknown
page, as the dataset does not contain completely identical layouts. 

We can also see that an important consistency principle for the nearest
neighbors holds: 
\begin{itemize}
\item Selecting a random page decreases the score.
\item Using a bigger search space for the nearest page increases the score.
\end{itemize}
\begin{table}[t]
\caption{\label{tab:Copypaste-baseline-results}Copypaste baselines results.
}

\begin{tabular*}{1\columnwidth}{@{\extracolsep{\fill}}|>{\raggedright}p{0.8\columnwidth}|>{\centering}p{0.1\columnwidth}|}
\hline 
\multirow{1}{0.8\columnwidth}{Experiments architecture\\
(and possible notable tweaks, see section \ref{sec:Experiments})} & Test micro\\
$F_{1}$ score\tabularnewline
\hline 
\hline 
Nearest page by embeddings and from validation set (standard) & 0.0582\tabularnewline
\hline 
Nearest page search from validation and train set  & 0.0599\tabularnewline
\hline 
Nearest page set to random  & 0.0552\tabularnewline
\hline 
\end{tabular*}
\end{table}

\paragraph{Oracle baseline}

In the \prettyref{tab:Copypaste-oracle} we can see the ``moderate
quality'' of the embeddings – only roughly $60\,\%$ of word-boxes
have their counterpart (class-wise) in the found nearest page.

When the nearest neighbor search is replaced with a completely random
pick, we can see an interesting property of the dataset – the number
of word-boxes that have a similar class in the random page increases
a little. This is because the distribution of class presence in the
pages is skewed. The reason for this is that vendors usually want
to incorporate more information into their business documents.

\begin{table}[t]
\caption{\label{tab:Copypaste-oracle}Oracle results. The metric ``Hits''
denotes the percentage of word-boxes that have their corresponding
class in the nearest page. }

\begin{tabular*}{1\columnwidth}{@{\extracolsep{\fill}}|>{\raggedright}p{0.8\columnwidth}|>{\centering}p{0.1\columnwidth}|}
\hline 
\multirow{1}{0.8\columnwidth}{Oracle setting} & Hits\tabularnewline
\hline 
\hline 
Nearest page by embeddings and from validation set (standard) & 59.52~\%\tabularnewline
\hline 
Nearest page search from validation and train set  & 60.43~\%\tabularnewline
\hline 
Nearest page set to random  & 60.84~\%\tabularnewline
\hline 
\end{tabular*}
\end{table}

\paragraph{Linear baseline}

The linear model has attained $0.3085$ test micro $F_{1}$ score.
Its performance justifies the progress from the basic Copypaste model
towards trainable architectures with similarity. But since it does
not beat the previous baseline results, it is proven that the similarity
principle alone does not help and thus justifies the design of more
complicated models.

\subsection{Results of architectures with similarity\label{subsec:Results}}

In this section we will look at all the designed architectures that
compete with the baselines.

The results for triplet loss architecture are presented in \prettyref{tab:Experimental-triplet},
the results for pairwise classification in \prettyref{tab:Experimental-pairwise}.

\begin{table}[t]
\caption{\label{tab:Experimental-triplet}Experimental results of triplet loss
architectures.}

\begin{tabular*}{1\columnwidth}{@{\extracolsep{\fill}}|>{\raggedright}p{0.8\columnwidth}|>{\centering}p{0.1\columnwidth}|}
\hline 
\multirow{1}{0.8\columnwidth}{Experiments architecture\\
(and possible notable tweaks, see section \ref{sec:Experiments})} & Test micro\\
$F_{1}$ score\tabularnewline
\hline 
\hline 
1x attention layer, loss-less variant  & 0.0619\tabularnewline
\hline 
2x attention layer, loss-less variant  & 0.0909\tabularnewline
\hline 
1x attention layer  & 0.1409\tabularnewline
\hline 
2x attention layer  & 0.1464\tabularnewline
\hline 
\end{tabular*}
\end{table}

\begin{table}[t]
\caption{\label{tab:Experimental-pairwise}Experimental results of pairwise
architectures.}

\begin{tabular*}{1\columnwidth}{@{\extracolsep{\fill}}|>{\raggedright}p{0.8\columnwidth}|>{\centering}p{0.1\columnwidth}|}
\hline 
\multirow{1}{0.8\columnwidth}{Experiments architecture\\
(and possible notable tweaks, see section \ref{sec:Experiments})} & Test micro\\
$F_{1}$ score\tabularnewline
\hline 
\hline 
\multicolumn{1}{|>{\raggedright}p{0.8\columnwidth}}{\center Pairwise classification} & \tabularnewline
\hline 
2x attention layer + refine section & 0.2080\tabularnewline
\hline 
2x attention layer  & 0.2658\tabularnewline
\hline 
1x attention layer & 0.2605\tabularnewline
\hline 
\end{tabular*}
\end{table}

Both pure triplet loss approaches and pairwise classification performed
better than simple Copypaste, but still worse than linear architecture.
Possible reasons could be:
\begin{itemize}
\item The existence and great prevalence of unclassified (uninteresting)
data in the documents. 

To this cause points the fact that all methods with hard-coded class
information (including simple linear baseline) scored better. Unfortunately,
this phenomena could be specific to the dataset. We could not replicate
the suboptimal results by modelling this situation in an existing
and otherwise successful task (omniglot challenge) by adding non-classifiable
types and by increasing the percentage of negative pairs.
\item Missing connections to the unknown page.

In the \prettyref{tab:Experimental-qa} we can see how the score drops
in QA architecture when we switch to the variant ``without query
all''. We conclude that even the best architecture needs a meaningful
information flow from the reference page itself and not only from
the nearest page. That information flow is missing in triplet loss
and pairwise classification.
\end{itemize}
To gain more insight, the architectures were tested on a different
target value, which was defined as ``does the text in these word-boxes
have the same length''. In this testing setting, the architectures
were able to achieve a significantly higher score of $0.7886$. This
supports our theory that the unclassified data (see above) was responsible
for the underperformance of triplet loss and pairwise classification,
since all data in the document was useful for the text lengths target.

\subsubsection{Query answer}

The query-answer architecture scored the best with a micro $F_{1}$
score of $0.9290$ with all the proposed architectural variants employed
at once. In \prettyref{tab:Experimental-qa} we present an ablation
study where we can see that each of the components (field of view,
query all, skip connection, filter, nearest search as defined in \prettyref{subsec:Query-answer-architecture})
related to QA architecture is needed, as the score drops if any single
one is turned off.

\begin{table}[t]
\caption{\label{tab:Experimental-qa}Experimental results of query answer architecture.}

\begin{tabular*}{1\columnwidth}{@{\extracolsep{\fill}}|>{\raggedright}p{0.8\columnwidth}|>{\centering}p{0.1\columnwidth}|}
\hline 
\multirow{1}{0.8\columnwidth}{Experiments architecture\\
(and possible notable tweaks, see section \ref{subsec:Query-answer-architecture})} & Test micro\\
$F_{1}$ score\tabularnewline
\hline 
\hline 
\textbf{All QA improvements in place } & \textbf{0.9290}\tabularnewline
\hline 
Fully anonymized dataset  & 0.7078\tabularnewline
\hline 
Only text features  & 0.8726\tabularnewline
\hline 
Nearest page set to random & 0.8555\tabularnewline
\hline 
Without field of view & 0.8957\tabularnewline
\hline 
Without query all  & 0.7997\tabularnewline
\hline 
Without skip connection  & 0.9002\tabularnewline
\hline 
Without filtering  & 0.8788\tabularnewline
\hline 
\end{tabular*}
\end{table}

Compared to the previous model (\prettyref{tab:Ex-Res-SImple-Model})
we get an improvement of $0.0825$ in the $F_{1}$ score. Also, the
experiment on the anonymized dataset and the dataset with only text
features shows the architecture is versatile enough to not fail the
task and to show similar improvement in the score on the anonymized
dataset (by $0.0950$). It also verifies that all of the visual, geometric
and textual features are important for good quality results.

\subsubsection{Qualitative comparison}

We will conclude with more qualitative analysis: comparing the best
Query Answer model and the Simple data extraction model. 

\begin{figure}
\includegraphics[width=1\columnwidth]{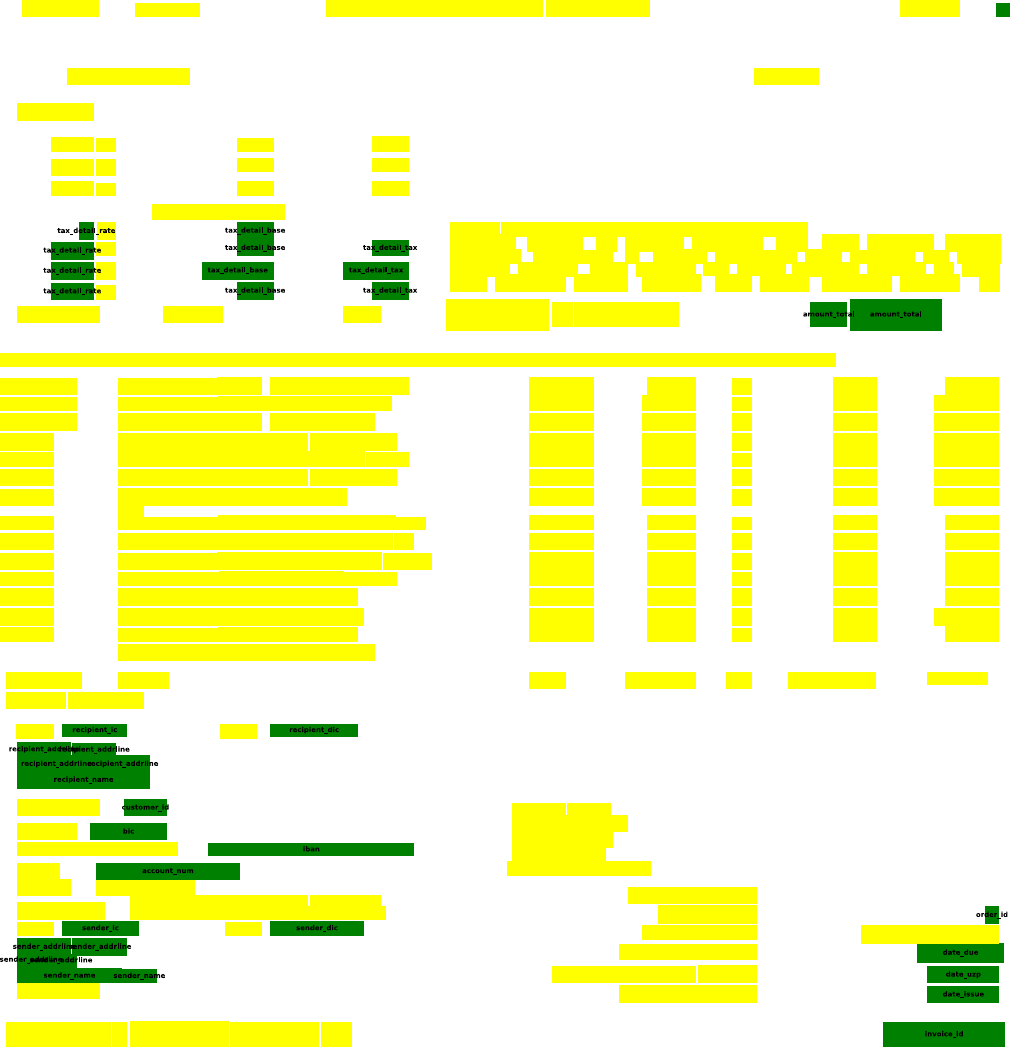}

\caption{\label{fig:QA-best}Best classification result of the Query answer
model – only true positives (green) and true negatives (yellow) can
be seen.}
\end{figure}

\begin{figure}
\includegraphics[width=1\columnwidth]{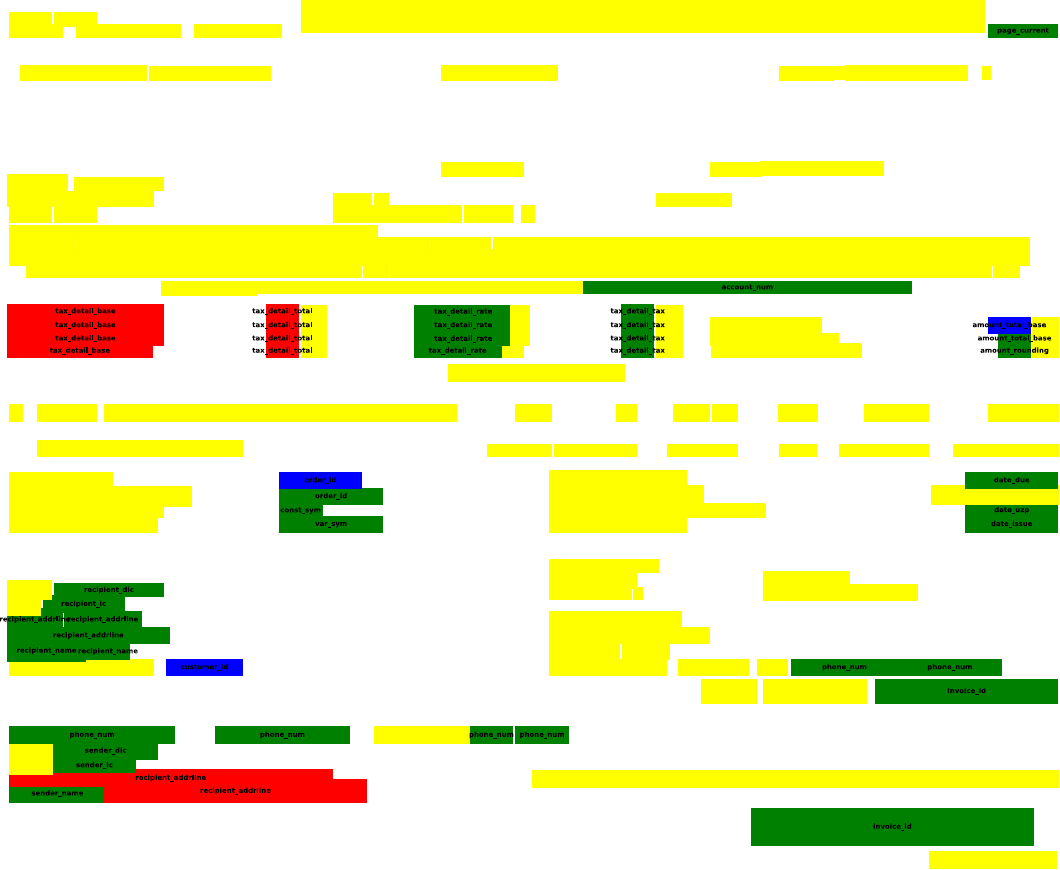}

\caption{\label{fig:QA-worst}Worst result of the query answer model. Each
blue and red area denotes a mistake.}
\end{figure}

\begin{figure}
\includegraphics[width=1\columnwidth]{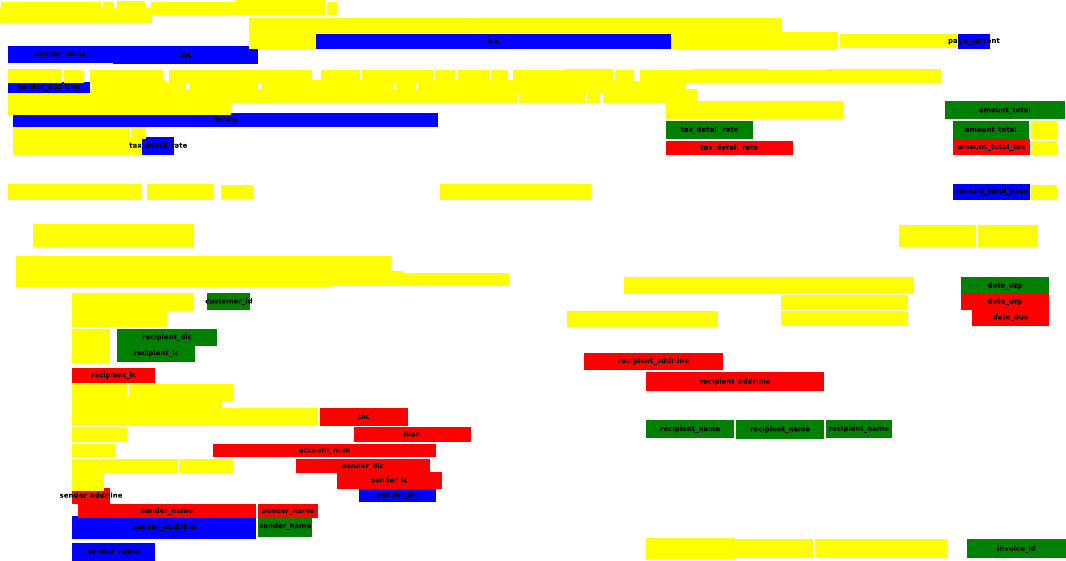}

\caption{\label{fig:Basleine-worst}Worst result of the Simple data extraction
model. Note the minimal count of true positive (green) areas and the
dominance of errors (blue and red).}
\end{figure}

First, to illustrate a manual inspection of prediction visualizations,
example visualizations in Figures \ref{fig:QA-best}, \ref{fig:QA-worst},
\ref{fig:Basleine-worst} are present, pages are selected from a random
subset of the test set. They show the best prediction from the Query
Answer model (\ref{fig:QA-best}), the worst prediction from the Query
Answer model (\ref{fig:QA-worst}) and finally the worst prediction
of the Simple data extraction model (\ref{fig:Basleine-worst}). The
green color indicates a successfully classified word-box (``true
positive''). The yellow color shows correctly classified unimportant
text (``true negative''). The blue color shows a misclassification
that should be yellow (an ``extra'') and the red color shows a misclassification
that should be green (a ``miss'').

Both the Simple data extraction model and the Query answer model have
examples of pages that look like results at \ref{fig:QA-best} and
are 100\% perfectly extracted (or classified). But the results vary
in the worst cases and that is why examples from both models are presented
in Figures \ref{fig:QA-worst} and \ref{fig:Basleine-worst}.

Motivated by this difference, we can look at which classes both models
extract best and worst. You can see those scores in the \prettyref{tab:Qual-Analyse}.

\begin{table}[t]
\caption{\label{tab:Qual-Analyse}Best and worst classes performance of Query
answer model and Simple data extraction model.}

\begin{tabular*}{1\columnwidth}{@{\extracolsep{\fill}}|>{\raggedright}p{0.5\columnwidth}|>{\centering}p{0.1\columnwidth}|>{\centering}p{0.1\columnwidth}|}
\hline 
\multirow{1}{0.5\columnwidth}{Best and worst performing fields\\
(and their scores)} & Simple – test\\
micro $F_{1}$ score & QA – test micro\\
$F_{1}$ score\tabularnewline
\hline 
\hline 
\multicolumn{1}{|>{\raggedright}p{0.5\columnwidth}}{Worst classes of Simple data extraction model} & \multicolumn{1}{>{\centering}p{0.1\columnwidth}}{} & \tabularnewline
\hline 
\raggedleft{}Page current & 0.30 & 0.90\tabularnewline
\hline 
\raggedleft{}Page total & 0.35 & 0.88\tabularnewline
\hline 
\raggedleft{}Terms & 0.62 & 0.78\tabularnewline
\hline 
\multicolumn{1}{|>{\raggedright}p{0.5\columnwidth}}{Best classes of Simple data extraction model} & \multicolumn{1}{>{\centering}p{0.1\columnwidth}}{} & \tabularnewline
\hline 
\raggedleft{}Recipient DIC & 0.94 & 0.96\tabularnewline
\hline 
\raggedleft{}Recipient IC & 0.94 & 0.97\tabularnewline
\hline 
\raggedleft{}Spec Symbol & 0.94 & 0.96\tabularnewline
\hline 
\multicolumn{1}{|>{\raggedright}p{0.5\columnwidth}}{Worst classes of Query answer} & \multicolumn{1}{>{\centering}p{0.1\columnwidth}}{} & \tabularnewline
\hline 
\raggedleft{}Order ID & 0.65 & 0.75\tabularnewline
\hline 
\raggedleft{}Terms & 0.62 & 0.78\tabularnewline
\hline 
\raggedleft{}Customer ID & 0.75 & 0.83\tabularnewline
\hline 
\multicolumn{1}{|>{\raggedright}p{0.5\columnwidth}}{Best classes of Query answer} & \multicolumn{1}{>{\centering}p{0.1\columnwidth}}{} & \tabularnewline
\hline 
\raggedleft{}Sender IC & 0.93 & 0.96\tabularnewline
\hline 
\raggedleft{}Spec Symbol & 0.94 & 0.96\tabularnewline
\hline 
\raggedleft{}Recipient IC & 0.94 & 0.97\tabularnewline
\hline 
\end{tabular*}
\end{table}

This detailed inspection shows that both models excel at classes that
usually appear together (but not in any fixed layout or order) in
business documents. Those classes are all the recipient information
(DIC, IC, Spec symbol) and the sender information. Moreover, recipient
information is usually required information on an invoice and thus
it is the most frequent class and so it is easy for the network to
excel at the detection thereof.

Interestingly, page numbering could be seen as an easy class to classify,
but is in reality classified with a very low score in the previous
model and jumps to a very high score when we switch to the QA model.
One possible reason for this is that the page number usually appears
alone somewhere near an edge of the page, and thus it’s nearest word-boxes
are random and might cause confusion for the GCN module and for convolution
over the sequence as well. When a similar page is presented to the
model, the score jumps higher possibly because the nearest page might
have page numbering in a similar position.

The QA model, as a candidate for an improvement from the previous
results, holds an important property we desire - we have verified
that the score for all classes has increased uniformly by at least
$0.02$ points (median gain being $0.04$), even for the previously
best performing classes. This property is important to verify, since
the QA architecture has the Simple data extraction model incorporated
internally, and we expect it to ``fall back'' to it when the nearest
page does not provide enough information. If this fallback would not
happen, some gradients would not be propagated correctly.

The improvement of some fields by only roughly $2\,\%$ may be seen
as a small improvement. But in reality (as stated in the introduction),
the $2\,\%$ improvement translates into less time and effort on more
than $500$ invoices per month. This reduced time and effort translates
to more than one thousand dollars of savings per month as well as
a reduction of carbon footprint.

\section{Conclusions\label{sec:Conclusions}}

Multiple baselines were provided and evaluated to gain more knowledge
about the data and ground the need for bigger and complicated models.

We have designed multiple ways to incorporate similarity and memory
– in terms of access to existing data – into the existing data extraction
model and studied the gains of various models in a fixed setting.
The successful gain was $8.25\,\%$ in $F_{1}$ score compared to
the previous results by using ``query-answer'' inspired architecture.
By the referenced heuristics, this improvement translates roughly
into $4000$ dollars savings of manual work per month for a middle-sized
company.

We have verified that all possible parts of the architecture are
needed in the training and prediction of the Query Answer model to
achieve the highest score. Moreover, the improvement holds even for
the case of the anonymized dataset.

In a qualitative analysis of the results, it was shown that the score
improvement is meaningful across all of the classes. Furthermore,
it was shown that the solution significantly boosted the previously
most-problematic classes.

For the other models that underperform (as triplet loss and pairwise
classification), we have identified the possible cause – being that
most words on the page do not belong to any class, and we have supported
the hypothesis with an additional experiment.

Further work could then incorporate a way to create some artificial
classes and measure the supposed increase in score for the triplet
and pairwise classification models. 

From the quantitative point of view, there is the opportunity to explore
and improve extraction scores by tuning all the possible parameters
of the system, namely the number of the nearest pages used and the
quality of the page embeddings. The page embeddings can be possibly
trained jointly with the word-box classificator. 

Qualitatively there are some possible new research questions:
\begin{itemize}
\item What is the effect of the size of the datasets? By exploring the effect
of the size of the training dataset and/or the search space for the
nearest pages, we could ask if (and when) the model needs to be retrained
and how does a sample of a difficult-to-extract document look like.
\item How to improve the means of generalization? Currently, the method
generalizes to unseen documents. In theory, we could desire a method
to generalize to new classes of words, since this way the model needs
to be retrained if a new class is desired to be detected and extracted.
\end{itemize}
In practice, our solution has one particular strength that does transform
these two points from potential blockers to just interesting research
questions. The model can fit into just one consumer-grade GPU and
trains from scratch for at most $4$ days using only one CPU process.
Compared to recent state-of-the-art NLP methods that take lots of
resources to train (such as \cite{brown2020language}), our model
can be retrained and/or fine-tuned for any particular use-case quickly
and effectively (even more with transfer learning techniques). Thus
any assorted problems are solved by the industrial standards \cite{burkov2020machine}.

As a part of this work, the dataset and source codes are published
in \cite{codeanddata} and should enable wider research of deep learning
models for information extraction, since – up until now – it was impossible
for researchers to collect a dataset of this size and quality.
\begin{acknowledgements}
The Rossum.ai team deserves thanks for providing the data and background
in which this work was able to grow.
\end{acknowledgements}

\section*{———————}

\bibliographystyle{spmpsci}
\addcontentsline{toc}{section}{\refname}\bibliography{SingleShotDocuments}

\end{document}